# Ground contact and reaction force sensing for linear policy control of quadruped robot

Harshita Mhaske
*California State University, Fullerton*
Fullerton, USA
harshie@csu.fullerton.edu

Aniket Mandhare
*California State University, Fullerton*
Fullerton, USA
aniketmmandhare@csu.fullerton.edu

Dr. Jidong Huang
*California State University, Fullerton*
Fullerton, USA
jhuang@fullerton.edu

Dr. Yu Bai
*California State University, Fullerton*
Fullerton, USA
ybai@fullerton.edu

*Abstract*— Designing robots capable of traversing uneven terrain and overcoming physical obstacles has been a longstanding challenge in the field of robotics. Walking robots show promise in this regard due to their agility, redundant DOFs and intermittent ground contact of locomoting appendages. However, the complexity of walking robots and their numerous DOFs make controlling them extremely difficult and computation heavy. Linear policies trained with reinforcement learning have been shown to perform adequately to enable quadrupedal walking, while being computationally light weight. The goal of this research is to study the effect of augmentation of observation space of a linear policy with newer state variables on performance of the policy. Since ground contact and reaction forces are the primary means of robot-environment interaction, they are essential state variables on which the linear policy must be informed. Experimental results show that augmenting the observation space with ground contact and reaction force data trains policies with better survivability, better stability against external disturbances and higher adaptability to untrained conditions.

*Keywords—quadruped, reinforcement learning, sensors*

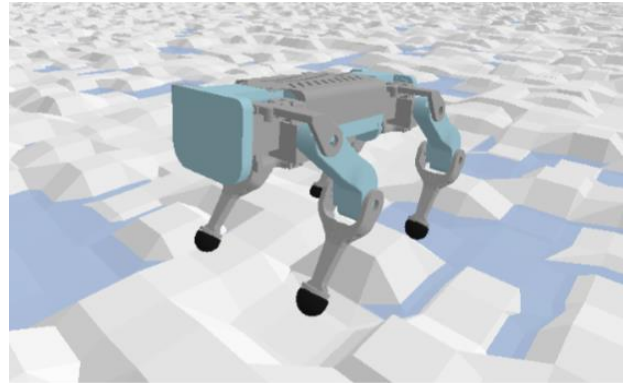

Fig.1: URDF of quadruped Minimal in simulation environment.

## I. Introduction

Unmanned ground vehicles (UGVs) find application in extreme, hazardous environments because they can be operated remotely for extended periods of time, reducing the risk to human operators. When deploying UGVs in unstructured environments, it is found that legged robots perform better at traversing uneven terrains than wheeled or tracked robots. Legged locomotion is enabled by discontinuous contact of locomoting appendages with the ground. This characteristic feature is useful in overcoming obstacles of varying geometry and can also be used for improving stability by controlling the location and force with which the appendages contact the ground. Ground reaction forces, which originate from the robot foot contacting the ground, are the primary means of physical interaction between the robot and its environment. As a result, accurate ground reaction force sensing [1] or estimation [2] and control is crucial for stable locomotion. Robots like StarlETH [3], HyQ [4], Mini Cheetah [5] implement control policies that demonstrate the importance of accurate ground reaction force estimation and control for dynamic robot behavior.

The control policies implemented by such robots can be computation heavy, increasing the compute power and thus the cost required for deployment on physical hardware. This makes the robot expensive, limiting access to hardware for many researchers. Hobbyist-level robots such as the Stanford Pupper [6] and SpotMicro [7] make use of FDM 3D printed or laser cut carbon fiber parts and RC servos, enabling a cost-friendly entry into legged robotics research. However, these robots are capable of simple, open-loop gaits only and cannot adapt to variations in the terrain. [8] presents the Policies Modulating Trajectory Generators approach, which modifies open-loop gaits based on feed-back from a 12-layer neural network to adapt to variations in the terrain. An inexpensive implementation of gait modulation has been demonstrated in [9], which uses a linear policy, trained off-line with reinforcement learning, to make modifications to a centrally generated gait trajectory, based on inertial measurements from an onboard IMU. As the linear policy is trained off-line in simulation, it can be deployed on a low-powered embedded microprocessor.

Both [8] and [9] use only inertial data as inputs to the modulating policies, and do not explicitly sense terrain variations. This paper explores the effects of using terrain data by way of ground contacts and ground reaction forces as inputs to the modulating policy. To that end, we develop a method to emulate ground reaction force and contact sensors in simulation. We augment the observation space of the policy described in [9] with these terrain data inputs, and train new policies based on IMU+Contacts and IMU+Force data. Finally, we evaluate the effects of observation space augmentation by first training the policies with the new inputs, followed by testing across various simulated scenarios viz. training performance, survivability in terms of distance travelled and time, stability against external disturbances and ability to traverse various slope gradients. The code implementation has been made open-source and is available at [10]

## II. Emulation of Ground Contact and Reaction Force Sensors

We used the Pybullet[11] environment for creating simulation of our robot. A URDF model of the robot is used for simulation, composed of numerous links with fixed or revolute joints between each link. The Pybullet API has a function to enable a force/torque sensor at a model joint. During the simulation, the forces and torques can be acquired by requesting the joint state. Apart from joint position and velocities, the joint state also includes reaction forces acting along and torques acting about constrained degrees of freedom of the joint.

Refer to fig.2 below. The robot model has a spherical foot at the distal end of each lower leg. A fixed joint is defined between the lower leg and the foot, with the lower leg being the parent link and the foot being the child link for the joint. The joint state data reports the forces in the joint reference frame, as applied by the parent link onto the joint. The sign of the reported forces is first flipped to get ground reaction force $fg$ in the joint frame.

The robot leg schematic in Fig. 3 shows the general layout of a single leg of our quadruped robot. The cylinders represent revolute joints connecting individual links, with cylinder axis being the axis of rotation. Each leg consists of 3 revolute joints: hip, shoulder and knee. The joint state data also includes information about the joint position along/about the freedom axis of a joint. For a revolute joint, joint position data constitutes the angular sweep of the joint. Using these functions, we emulate a set of joint angle sensors that fetch the angular joint positions for 12 joints of the robot at each step of the simulation.

Joint angles acquired at each simulation step are used to transform $fg$ in joint frame $fg_j$ to $fg$ in the robot base frame $fg_R$ by application of rotation transforms. Let ${}^aT_b$ represent a transform from frame b to frame a. The joint frame to robot frame transform is calculated as:

$${}^RT_j = {}^RT_H \, {}^HT_{UL} \, {}^{UL}T_{LL} \, {}^{LL}T_j \qquad (1)$$

where j, LL, UL, H and R represent the reference frames of the fixed foot joint, lower leg, upper leg, hip and the robot base links respectively. Thus, it follows from (1) that,

$$F_l = fg_R = {}^RT_j \, fg_j \qquad (2)$$

where $F_l$ is the ground reaction force from a leg $l$ in the robot base frame. The legs of the quadruped are designated as front left (FL), front right (FR), back left (BL) and back right (BR) so that $l \in \{FL, FR, BL, BR\}$. To emulate a contact sensor, we define $C_l$ as a threshold function of $F_l$ so that,

$$C_l = \lambda(F_l) = \begin{cases} 1, & F_l > F_{Threshold} \\ 0, & F_l \leq F_{Threshold} \end{cases} \qquad (3)$$

where $F_{Threshold}$ is a small positive value.

### III. VERIFICATION OF GROUND REACTION FORCE SENSORS

We verified the ground reaction forces from the feet at stance position, with the robot standing on flat ground and on a 5° inclined plane in simulation. We collected the GRF values at a random time-step while the robot was stationary (refer table I). The ground reaction forces collected have been tabulated below, and we calculate the resultant force acting on the robot in the robot base frame. The robot weight in simulation is

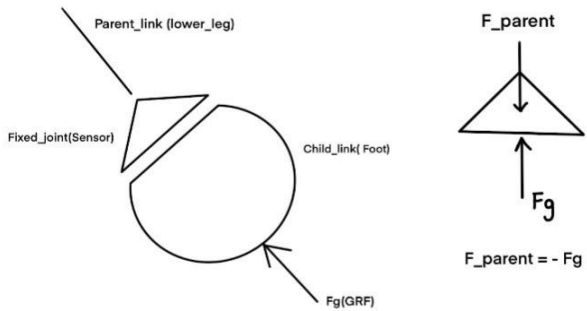

Fig. 2: Ground reaction forces acting on the foot

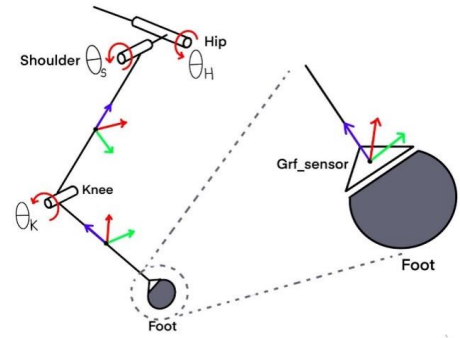

Fig 3. Robot leg schematic with reference frames

28.6N. In both cases, the total force on the robot matches the robot weight, acting in a direction opposite to gravity to support the robot. This means that the force values collected are correct within error limits.

Table I.: GRFs for robot standing on flat ground

| Foot | Ground reaction force components in N in R frame | | |
|---|---|---|---|
|  | $f_x$ | $f_y$ | $f_z$ |
| FL | 2.461 | 2.004 | 6.075 |
| FR | 0.256 | -3.385 | 8.990 |
| BL | -4.229 | 3.778 | 8.445 |
| BR | 1.516 | -2.396 | 6.101 |
| Total | $F_x$ = 0.004 | $F_y$ = 0.001 | $F_z$ = 29.611 |
| Resultant F= 29.611 N; $\theta$ = $\tan^{-1}(F_x/F_z)$ = 0.007° | | | |

Table II.: GRFs for robot standing on 5° incline

| Foot | Ground reaction force components in N in R frame | | |
|---|---|---|---|
|  | $f_x$ | $f_y$ | $f_z$ |
| FL | 1.401 | 1.472 | 6.046 |
| FR | 1.398 | -1.691 | 7.027 |
| BL | -0.43 | 1.493 | 8.416 |
| BR | 0.224 | -1.240 | 8.030 |
| Total | $F_x$ = 2.593 | $F_y$ = 0.034 | $F_z$ = 29.519 |
| Resultant F= 29.63 N; $\theta$ = $\tan^{-1}(F_x/F_z)$ = 5.020° | | | |

### IV. POLICY

The policy from [9] is a linear map of $n$ observations to $m$ desirable actions. Thus, the policy itself is $n \times m$ matrix of scalar weights. Observation space $o_t$ for the IMU only policies is composed of roll $r$ and pitch $p$ angles, angular velocities $\omega$ and angular accelerations $v$ of the robot base link, and the gait phase $S_\ell(t)$ of each foot. The policy that uses these observation serves as a baseline against which we compare policies that use ground reaction forces $f_G$ and foot contact data C as additional observations. So, our policies are differentiated as:

1) IMU: with $o_t = [r,p,\omega,v,S(t)]^\top \in R^{12}$
2) IMU+Force: with $o_t = [r,p,\omega,v,S(t), f_G]^\top \in R^{24}$
3) IMU+Contacts: with $o_t = [r,p,\omega,v,S(t), C]^\top \in R^{16}$

The action space is composed of gait parameters viz. nominal clearance height and ground penetration depth for the Bezier curve trajectory generator, and translational displacements $\Delta x, \Delta y, \Delta z$ for each foot. The final foot position is a vector

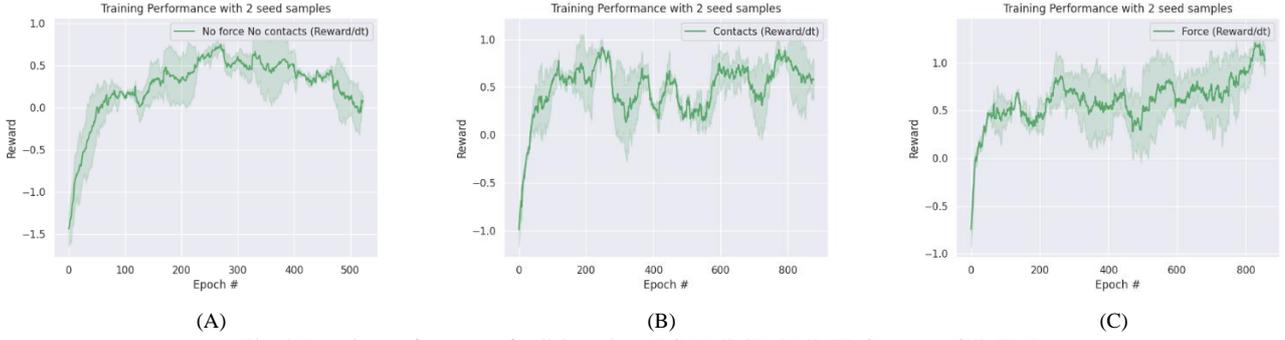

Fig. 4: Learning performance of policies using ARS (A) IMU; (B) IMU+Contacts; (C)IMU+Force

sum of the co-ordinates commanded by the Bezier gait generator and the translational displacements.

Augmented Random Search (ARS) [12] is a policy optimization technique that utilizes random parameter search for reinforcement learning. In each ARS optimization step, we conduct 16 rollouts per iteration, applying a learning rate of 0.03 and an exploration noise of 0.05 for parameter updates. Specifically, each of the 16 rollouts samples a new parameter, denoted as $\theta = \theta + \Delta\theta$, where $\Delta\theta$ follows a normal distribution N(0, 0.05). Here, N represents a normal distribution with a mean of 0 in $R^{12 \times 14}$ and a variance of 0.05.

Each training episode runs for T = 5000 steps (equivalent to 50 seconds). Between each episode, the robot dynamics and terrain are randomized to train a robust policy. The reward function $r_t$ is defined as:

$$r_t = \Delta x - 10\,(|r| + |p|) - 0.03 \sum |\omega| \qquad (4)$$

where $\Delta x$ represents the horizontal distance travelled by the robot in one time step. We observed that normalizing the final episode reward by the total number of steps enhances policy learning. This adjustment reduces the penalty for abrupt failures following an otherwise successful run, ultimately promoting longer survival times.

## V. RESULTS

We present several experiments to evaluate our approach for improving legged locomotion with a simple linear policy.

### A. Training performance

We collected the reward per time step for each epoch of the training for all three policies. Fig. 4 shows the reward growth for each policy over the course of training. The reward per time step after the first epoch is the highest for IMU+Force based policy, followed by IMU+Contacts, with IMU being the lowest. IMU based policy shows a downward trend in reward growth across epochs. For IMU+Contact based policy, the reward growth shows a trend to oscillate between 0 and 1, and does not show an upward trend. IMU+Force policies show a stable reward growth across epoch, and we see a converging upward trend.

### B. Survival data

We extracted survival data for IMU, IMU+Contacts and IMU+Force observation space policies, trained over a 0.104m maximum terrain height. The survival data consists of episode time, position reached, and rewards accumulated per episode.

Episodes that were terminated when the robot travelled a distance of 100m or at a maximum duration of 50000 timesteps are termed as "alive". If the robot falls over before the maximum duration, that episode is termed as "dead". Policies that remain alive but do not cover a large distance remain stuck at a spot in the terrain, without falling. The plots below show the survival data for policies deployed over a terrain with maximum height 0.104 m (~55% of robot height) in terms of the distance and time survived. The collected data was sorted in ascending order for clarity. From fig. 5, we see that the number of episodes for which the robot walks the full 100m is the largest for IMU+Contacts, followed by IMU+Force and IMU policies. The plot in fig. 6 shows that the number of policies lasting the longest duration is largest for IMU+Force policies, followed by IMU+Contacts and IMU.

Table III: Survival Data

| Distance | 0 to 5 m | | 5 to 90 m | | >90m | |
|---|---|---|---|---|---|---|
| Policy | *dead* | *alive* | *dead* | *alive* | *dead* | *alive* |
| IMU | 352 | 15 | 276 | 9 | 0 | 348 |
| IMU+Force | 194 | 66 | 207 | 123 | 0 | 410 |
| IMU+C | 281 | 18 | 199 | 12 | 0 | 490 |

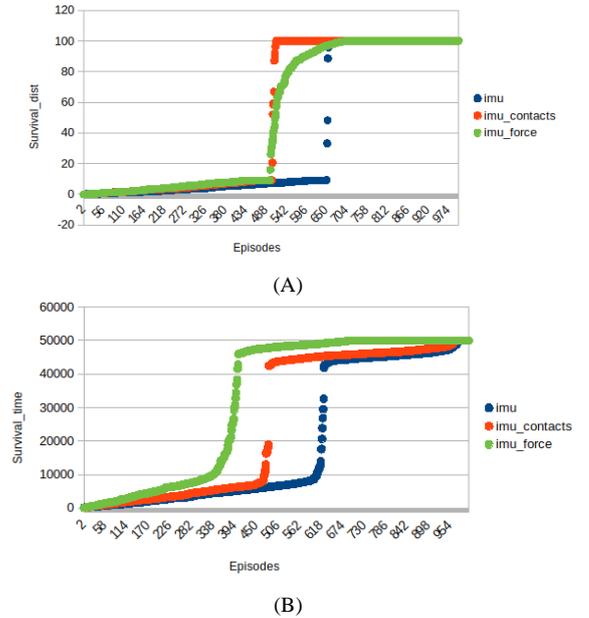

Fig. 5: Episode survival distance (A) and time (B) for policies deployed over 0.104m max terrain height

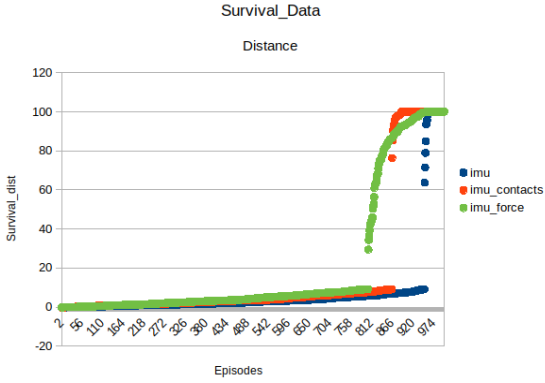

Fig.6: Survival distance over untrained terrain height

Policies last longer for two reasons: The robot hasn't fallen over yet, or the robot hasn't walked the full 100m. We can draw the following inferences:

1) IMU+Force and IMU+contact policies outperform IMU policies in both survival time and distance.

2) IMU+Force policies are slower at covering the distance, but less likely to fall over as compared to IMU+Contact policies. By removing the episode time restriction, we would see a larger number of IMU+force policies completing the full 100m.

3) IMU+Contact policies are unlikely to get stuck in the ueven terrain, as evidenced by the relatively smaller number of IMU+Contact policies in the 20m to 80m range in fig. 5(A) and 10000 to 40000 time-steps survival range in fig. 5(B).

*C. Adaptability over higher terrain.*

We trained all three types of policies over a 0.104m maximum terrain height. We then deployed the same policies over an increased terrain height of 0.128m to check which set of observations would enable the policy to better adapt to untrained conditions. From fig.6, we see the number of surviving contact-based policies reduce considerably. However, the number of policies  completing 100m of walking still follows the trend IMU+Contact>IMU+Force>IMU.

*D. Stability over flat terrain and against disturbances*

To establish a baseline stability behavior, we deployed all three policies on flat ground. From simulation of each policy, we collect the roll and pitch angle values of the robot base orientation for each simulation time-step. Referring to fig.7, we can see that IMU+Force observation-based policy has the lowest overall roll magnitudes while walking undisturbed on flat ground.

Next, we tested policy performance against external disturbances. The simulation has the robot walking on flat ground. We then launch a ball of mass 0.5 kg at a velocity of 3.5 m/s in the y-direction at the robot, while it is walking in the x-direction, to make it fall over. From fig.8 and 9, we can see that the peak angle values follow the trend IMU>IMU+Contact>IMU+Force

*E. Survival and agility over slopes*

Next, we tested the performance of policies on a 8 degrees upward slope and downward slope. The 8° upward incline starts at a distance of 1 m from starting point. The decline starts at a distance of 1.5m, and transitions into flat ground at 2m. Episodes that were terminated when the robot travelled a distance of 3m (having successfully traversed the slope) or at a maximum duration of 50000 timesteps are termed as "alive". If the robot falls over before the maximum duration, that episode is termed as "dead". Each policy was tested for 50 episodes. From table IV, we can see that IMU+Contact policies outperform both IMU+Force and IMU based policies with a 100% success rate.

## VI. ENABLING CONTACT SENSING ON REAL ROBOT

Simulation testing of force-based and contact-based policies show a clear advantage of terrain sensing via contacts over ground reaction forces. So, we chose to implement foot contact detection on the physical robot hardware. Keeping in mind the goal of minimizing cost, we explored various methods by which such contact estimation could be enabled.

Table IV: Survivability over 8° slope

| Distance | 0 to 1 m | | 1 to 2 m | | >2 m | |
|---|---|---|---|---|---|---|
| Policy | *dead* | *alive* | *dead* | *alive* | *dead* | *Alive* |
| IMU | 9 | 4 | 13 | 0 | 0 | 24 |
| IMU+Force | 11 | 0 | 9 | 0 | 0 | 30 |
| IMU+C | 0 | 0 | 0 | 0 | 0 | 50 |

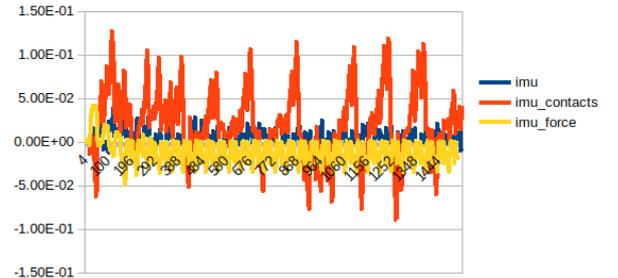

Fig.7: Roll during flat terrain walking

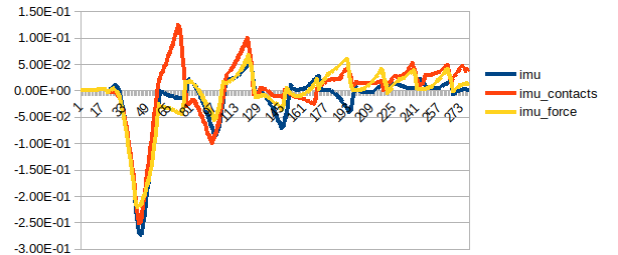

Fig.8: Roll disturbance and recovery

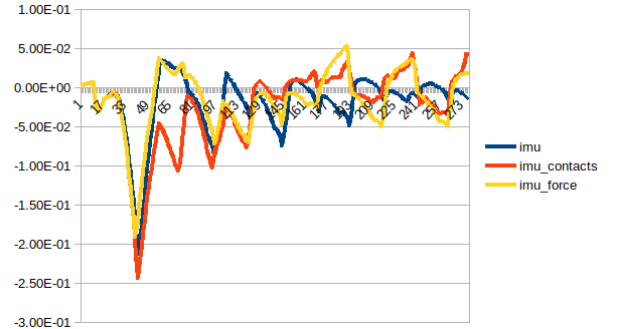

Fig.9: Pitch disturbance and recovery

Force sensitive resistors (FSR) are thin-film, low-cost devices, that show a decrease in resistance when a force is applied on them. We chose an FSR with a small form factor (10mm X 20 mm) and a force sensing range of 0.2N to 20N. The FSR is sandwiched between the lower-leg and foot of the robot. The foot has an integrated cylindrical boss that acts on the active region of the FSR (fig.10). The lower leg and foot are separated by 4 silicon O-rings when assembled. When the foot contacts the ground, the O-rings are compressed and the cylindrical boss of the foot applies force on the FSR, decreasing its resistance. When the foot breaks contact with the ground, the O-rings decompress, causing the foot to return to its original position. The force applied on the FSR is decreased, increasing its resistance.

We use a voltage divider circuit to convert the varying resistance value to a voltage value, which is then read by a microcontroller. This voltage reading is input into a threshold step-function like (3) for contact detection. We constructed a physical prototype of the contact-detecting foot and assembled it onto the back right leg of the robot. To verify the contact detection, we built the circuit described above. We used an LED to visually display the state of contact, so that the LED is ON when the back right foot contacts the ground and OFF when the foot is in the air. The successful demonstration of contact detection is shown in fig.11, and is available in video format at [10].

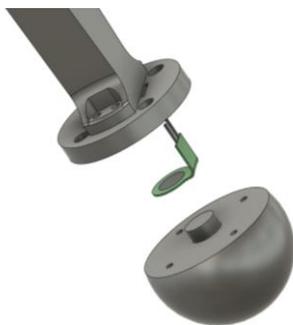

Fig. 10: Exploded view of the robot leg

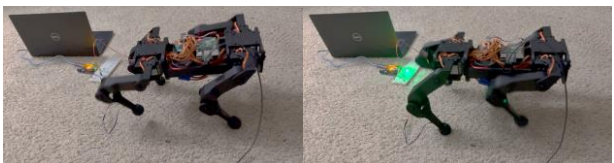

Fig. 11: Demonstrating contact sensing on physical hardware

## VII. CONCLUSION

Comparison of the reward growth of different policies across episodes during training, we can see that IMU+Force policies show a more rapid and stable reward growth. Thus, policy learning by ARS can be improved using a larger number of richer data inputs.

When terrain sensing data is included in the observation space of a policy, either by the way of contacts or ground reaction forces, the survivability, stability and adaptability of the robot shows significant improvement. Between the two methods, force-based sensing shows lower roll and pitch oscillations, while contact based policies are less likely to get stuck at a point in the terrain. For linear policies modulating a baseline foot trajectory, terrain sensing by the way of contacts enables better robot survivability across different terrains.

## VIII. FUTURE WORK

We are interested in implementing the tested policies on physical hardware. For ground reaction force sensing, we will investigate the construction of low cost and compact three-axis sensor, as shown in [13] and [14]. We have already demonstrated the working of an integrated contact sensor for one foot, and will be extending it to all four feet, similar to [15]. Experimental setups for comparative testing of the robot's hardware will also be designed, to benchmark robot performance.